\documentclass[runningheads]{llncs}
\usepackage{graphicx}

\usepackage{tikz}
\usepackage{comment}
\usepackage{amsmath,amssymb} 
\usepackage{color}
\usepackage{xspace}
\usepackage{tabulary}
\usepackage{url}

\newcolumntype{x}[1]{>{\centering\arraybackslash}p{#1pt}}

\newcommand{\csmall}{\fontsize{8}{9.5}\selectfont}

\makeatletter
\DeclareRobustCommand\onedot{\futurelet\@let@token\@onedot}
\def\@onedot{\ifx\@let@token.\else.\null\fi\xspace}

\def\eg{\emph{e.g}\onedot}

\def\etal{\emph{et al}\onedot}
\makeatother

\usepackage[accsupp]{axessibility}  

\begin{document}

\pagestyle{headings}
\mainmatter
\def\ECCVSubNumber{6129}  

\title{GraphCSPN: Geometry-Aware Depth Completion via Dynamic GCNs} 

\titlerunning{GraphCSPN: Geometry-Aware Depth Completion via Dynamic GCNs}
%

\author{Xin Liu\inst{1} \and
Xiaofei Shao\inst{2} \and
Bo Wang\inst{2} \and
Yali Li\inst{1} \and
Shengjin Wang\inst{1}\thanks{Corresponding author}}
\authorrunning{Liu et al.}
%

\institute{Beijing National Research Center for Information Science and Technology (BNRist)\\
Department of Electronic Engineering, Tsinghua University
\email{xinliu20@mails.tsinghua.edu.cn,  \{liyali13, wgsgj\}@tsinghua.edu.cn}
\and
Deptrum Ltd. \\
\email{\{xiaofei.shao, bo.wang\}@deptrum.com}
}
\maketitle

\begin{abstract}
Image guided depth completion aims to recover per-pixel dense depth maps from sparse depth measurements with the help of aligned color images, which has a wide range of applications from robotics to autonomous driving. However, the 3D nature of sparse-to-dense depth completion has not been fully explored by previous methods. In this work, we propose a \textbf{Graph} \textbf{C}onvolution based \textbf{S}patial \textbf{P}ropagation \textbf{N}etwork (\textbf{GraphCSPN}) as a general approach for depth completion. First, unlike previous methods, we leverage convolution neural networks as well as graph neural networks in a complementary way for geometric representation learning. In addition, the proposed networks explicitly incorporate learnable geometric constraints to regularize the propagation process performed in three-dimensional space rather than in two-dimensional plane. Furthermore, we construct the graph utilizing sequences of feature patches, and update it dynamically with an edge attention module during propagation, so as to better capture both the local neighboring features and global relationships over long distance. Extensive experiments on both indoor NYU-Depth-v2 and outdoor KITTI datasets demonstrate that our method achieves the state-of-the-art performance, especially when compared in the case of using only a few propagation steps. Code and models are available at the project page \footnote{\url{https://github.com/xinliu20/GraphCSPN_ECCV2022}}.
\keywords{Depth completion, Graph neural network, Spatial propagation}
\end{abstract}

\section{Introduction}

Depth perception plays an important role in various real-world applications of computer vision, such as navigation of robotics~\cite{el2012study,jing2017comparison} and autonomous vehicles~\cite{bai2020depthnet,farahnakian2020rgb}, augmented reality~\cite{diaz2017designing,du2020depthlab}, and 3D face recognition~\cite{gordon1992face,pan20053d}. However, it is difficult to directly acquire dense depth maps using depth sensors, including LiDAR, time-of-flight or structure-light-based 3D cameras, either because of the inherent limitations of hardware or due to the interference of surrounding environment. Since depth sensors can only provide sparse depth measurements of object at distance, there has been a growing interest within both academia and industry in reconstructing depth in full resolution with the guidance of corresponding color images. 

\begin{figure}[t]
\centering
\begin{minipage}{\textwidth}
\newcommand{\figWidth}{ 0.165\linewidth } 
\setlength\tabcolsep{0.1mm} 
\begin{tabular}{ c c c c c c}
  \begin{minipage}[m]{\figWidth}\centering
  \includegraphics[width=\linewidth]{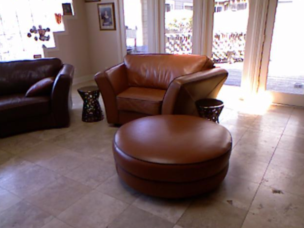} \\
  \tiny (a) RGB
  \end{minipage}
  & 
  \begin{minipage}[m]{\figWidth}\centering
  \includegraphics[width=\linewidth]{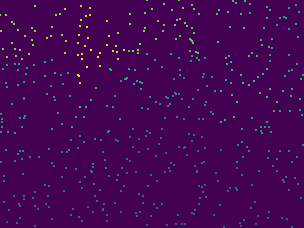} \\
  \tiny (b) sparse depth
  \end{minipage}
  &
  \begin{minipage}[m]{\figWidth}\centering
  \includegraphics[width=\linewidth]{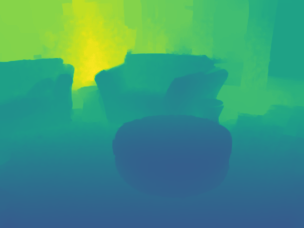} \\
  \tiny (c) ground truth
  \end{minipage}
  &
  \begin{minipage}[m]{\figWidth}\centering
  \includegraphics[width=\linewidth]{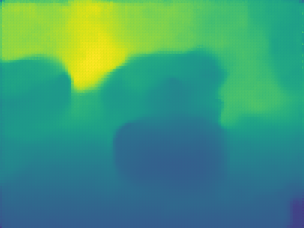} \\
  \tiny (d) initial depth
  \end{minipage}
  & 
  \begin{minipage}[m]{\figWidth}\centering
  \includegraphics[width=\linewidth]{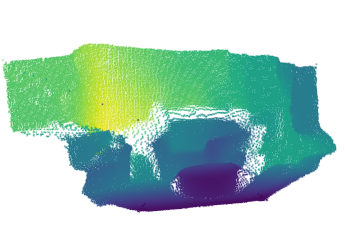} \\
  \tiny (e) propagation 
  \end{minipage}
    &
  \begin{minipage}[m]{\figWidth}\centering
  \includegraphics[width=\linewidth]{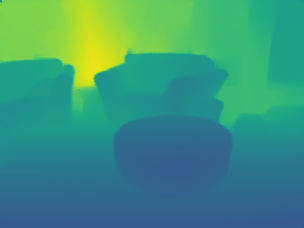} \\
  \tiny (f) final prediction
  \end{minipage}
\end{tabular}
\end{minipage}
\caption{{\bf Illustration of depth completion task using our framework.} A backbone model receives the sparse depth map and corresponding RGB image as input and outputs an initial depth prediction. And then the initial depth is iteratively refined by our geometry-aware GraphCSPN in 3D space to produce the final depth prediction. Sparse depth map (b) has less than 1\% valid values and is dilated for visualization.}
\label{fig:RGBd}
\end{figure}

To address this challenging problem of sparse-to-dense depth completion, a wide variety of methods have been proposed. Early approaches~\cite{wang2008stereoscopic,shen2013layer,park2014high} mainly focus on handcrafted features which often lead to inaccurate results and have poor generalization ability. Recent advance in deep convolutional neural networks (CNN) has demonstrated its promising performance on the task of depth completion~\cite{cheng2018depth,park2020non,hu2021penet}. Although CNN based methods have already achieved impressive results for depth completion, the inherent local connection property of CNN makes it difficult to work on depth map with sparse and irregular distribution, and hence fail to capture 3D geometric features. Inspired by graph neural networks (GNN) that can operate on irregular data represented by a graph, we propose a geometry-aware and dynamically constructed GNN. And it is combined with CNN in a complementary way for geometric representation learning, in order to fully explore the 3D nature of depth prediction.

Among the state-of-the-art methods for depth completion, spatial propagation~\cite{liu2017learning} based models achieve better results and are more efficient and interpretable than direct depth completion models~\cite{ma2018sparse}. Convolutional spatial propagation network (CSPN)~\cite{cheng2018depth} and other methods built on it~\cite{cheng2020cspn++,park2020non} learn the initial depth prediction and affinity matrix for neighboring pixels, and then iteratively refine the depth prediction through recurrent convolutional operations. Recently, Park \etal ~\cite{park2020non} propose a non-local spatial propagation network (NLSPN) which alleviates the mixed-depth problem on object boundaries. Nevertheless, there are several limitations regarding to such approaches. Firstly, the neighbors and affinity matrix are both fixed during the entire iterative propagation process, which may lead to incorrect predictions because of the propagation of errors in refinement module. In addition, the previous spatial propagation based methods perform propagation in two dimensional plane without geometric constraints, neglecting the 3D nature of depth estimation. Moreover, they suffer from the problem of demanding numerous steps (\eg, 24) of iteration to obtain accurate results. The long iteration process indicates the inefficiency of information propagation and may limit their real-world applications. 

To address the limitations stated above, we relax those restrictions and generalize all previous spatial propagation based methods into a unified framework leveraging graph neural networks. The motivation behind our proposed model is not only because GNN is capable of working on irregular data in 3D space, but also the message passing principle~\cite{gilmer2017neural} of GNN is strongly in accord with the process of spatial propagation. We adopt an encoder-decoder architecture as a simple while effective multi-modality fusion strategy to learn the joint representation of RGB and depth images, which is utilized to construct the graph. Then the graph propagation is performed in 3D space under learnable geometric constraints with neighbors updated dynamically for every step. Furthermore, to facilitate the propagation process, we propose an edge attention module to aggregate information from corresponding position of neighboring patches. In summary, the main contributions of the paper are as follows:
\begin{itemize}
\item We propose a graph convolution based spatial propagation network for sparse-to-dense depth completion. It is a generic and propagation-efficient framework and only requires 3 or less propagation steps compared with 18 or more steps used in previous methods.
\item We develop a geometry-aware and dynamically constructed graph neural network with an edge attention module. The proposed model provides new insights on how GNN can help to deal with 2D images in 3D perception related tasks.
\item Extensive experiments are conducted on both indoor NYU-Depth-v2 and outdoor KITTI datasets which show that our method achieves better results than previous state-of-the-art approaches. 
\end{itemize}

\section{Related Work}

{\bf Depth Completion} Image guided depth completion is an important subfield of depth estimation, which aims to predict dense depth maps from various input information with different modalities. However, depth estimation from only a single RGB image often leads to unreliable results due to the inherent ambiguity of depth prediction from images. To attain a robust and accurate estimation, Ma and Karaman~\cite{ma2018sparse} proposed a deep regression model for depth completion, which boosts the accuracy of prediction by a large margin compared to using only RGB images. To address the problems of image guided depth completion, various deep learning based methods have been proposed – \eg, sparse invariant convolution~\cite{hua2018normalized,eldesokey2018propagating,huang2019hms}, confidence propagation~\cite{eldesokey2019confidence,hekmatian2019conf}, multi-modality fusion~\cite{tang2020learning,hu2021towards}, utilizing Bayesian networks~\cite{Qu_2021_ICCV} and unsupervised learning~\cite{Wong_2021_ICCV}, exploiting semantic segmentation~\cite{jaritz2018sparse,schneider2016semantically} and surface normal~\cite{qiu2019deeplidar,xu2019depth} as auxiliary tasks.

\noindent{\bf Spatial Propagation Network} The spatial propagation network (SPN) proposed in~\cite{liu2017learning} can learn semantically-aware affinity matrix for vision tasks including depth completion. The propagation of SPN is performed sequentially in a row-wise and column-wise manner with a three-way connection, which can only capture limited local features in an inefficient way. Cheng \etal ~\cite{cheng2018depth} applied SPN on the task of depth completion and proposed a convolutional spatial propagation network (CSPN), which performs propagation with a manner of recurrent convolutional operation and alleviates the inefficiency problem of SPN. Later, CSPN++~\cite{cheng2020cspn++} was proposed to learn context aware and resource aware convolutional spatial propagation networks and improves the accuracy and efficiency of depth completion. Recently, Park \etal ~\cite{park2020non} proposed NLSPN to learn deformable kernels for propagation which is robust to mixed-depth problem on depth boundaries. Following this family of approaches based on spatial propagation, we further propose a graph convolution based spatial propagation network (GraphCSPN) which provides a generic framework for depth completion. Unlike previous methods, GraphCSPN is constructed dynamically by learned patch-wise affinities and performs efficient propagation with geometrically relevant neighbors in three-dimensional space rather than in two-dimensional plane.

\noindent{\bf Graph Neural Network} Graph neural networks (GNNs) receive a set of nodes as input and are invariant to permutations of the node sequence. GNNs work directly on graph-based data and capture dependency of objects via message passing between nodes~\cite{zhou2018graph,gilmer2017neural,li2021deepgcns_pami}. GNNs have been applied in various vision tasks, such as image classification~\cite{garcia2017few,wang2018zero}, object detection~\cite{hu2018relation,gu2018learning} and visual question answering~\cite{teney2017graph,narasimhan2018out}. Unlike previous depth completion methods using GNNs for multi-modality fusion~\cite{zhao2020adaptive}, learning dynamic kernel~\cite{xiong2020sparse}, we leverage GNNs as its message passing principle is in accord with spatial propagation. In addition, we develop a geometry-aware and dynamically constructed GCN with edge attention to aggregate and update information from neighboring nodes.

\section{Method}
In this section, we start by introducing the spatial propagation network (SPN) and previous methods that build on SPN. To address the limitations of those methods, we present our graph convolution based spatial propagation network and show how it extends and generalizes earlier approaches into a unified framework. Then we describe in details every component of the proposed framework, including graph construction, neighborhood estimation and graph propagation. Furthermore, a theoretical analysis of our method from the perspective of anisotropic diffusion is provided in supplementary material.
\subsection{Spatial Propagation Network}
In the task of sparse-to-dense depth completion, spatial propagation network~\cite{liu2017learning} is designed to be a refine module working on the initial depth prediction in a recursive manner. The initial depth prediction can be the output of an encoder-decoder network or other networks utilizing more complicated multi-modality fusion strategies. After several iteration steps, the final prediction result is obtained with more detailed and accurate structure. We formulate the updating process of previous methods~\cite{liu2017learning,cheng2018depth,cheng2020cspn++,park2020non} in each propagation step as follows: 
\begin{equation}
    d^{s+1}_{p,q} {=} \mu(\mathbf{D}^{s}|\mathbf{A}_{p,q},\mathcal{N}_{p,q}) {=} a_{p,q}^{p,q}d^{s}_{p,q} {+} \sum_{(i,j) \in \mathcal{N}_{p,q}}a_{p,q}^{i,j}d^{s}_{i,j},
\label{eq:update}
\end{equation}
where $s$ denotes the iteration step; $d_{p,q} \in \mathbf{D}$ is the depth value at coordinate $(p, q)$ of 2D depth map; $(i,j) \in \mathcal{N}_{p,q}$ indicates the coordinate of neighbors at $(p, q)$; and $a_{p,q}^{i,j} \in \mathbf{A}_{p,q}$ represents the affinity between the pixels at $(p, q)$ and $(i, j)$. During the spatial propagation, each depth value is updated by its neighbors according to the affinity matrix which defines how much information should be passed between neighboring pixels. Previous methods based on SPN construct the neighborhood through simple coordinate shift, which can be summarized as:
\begin{equation}
    \mathcal{N}_{p,q} {=} \{(p+m,q+n)|(m, n) \in \Gamma(\mathbf{D}| p, q), (m,n) \neq (0, 0)\},
\label{eq:neighbor}
\end{equation}
where $(m, n)$ represents the coordinate shift of neighbors; and $\Gamma$ denotes the estimation function of neighbors given the initial depth prediction $\mathbf{D}$ and position $(p, q)$ as input. For local spatial propagation methods, the neighbor estimation function $\Gamma$ is irrelevant to the input and falls to the fixed coordinate set, \eg $\left\{-1, 0, 1\right\}$ for $3\times3$ kernel. The original SPN~\cite{liu2017learning} performs propagation sequentially in a row-wise and column-wise manner with three-way connections which is inefficient. CSPN~\cite{cheng2018depth} updates the depth estimation by making use of all local neighbors simultaneously. CSPN++~\cite{cheng2020cspn++} is able to learn square kernel of different sizes. The neighborhood estimation function of NLSPN~\cite{park2020non} can learn non-local neighbors according to the pixel positions in 2D plane. 

\begin{figure}[t]
\centerline{\includegraphics[width=\linewidth]{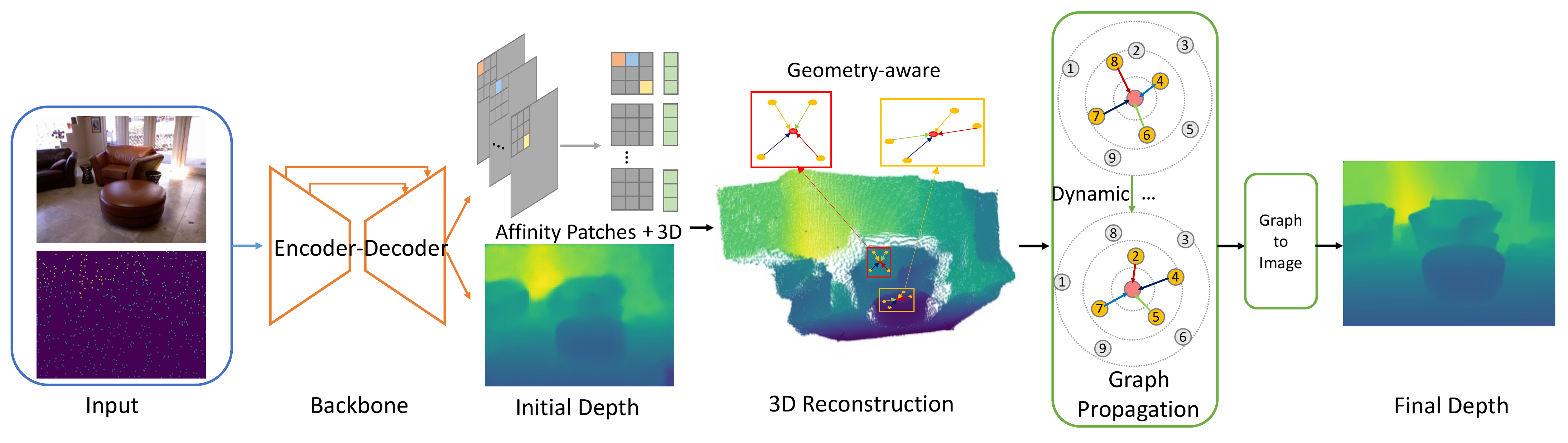}}
\caption{ \textbf{Overview of the proposed network architecture} (Best viewed in color). There are mainly two parts of the network. The first part utilizes an encoder-decoder to jointly learn the initial depth map and affinity matrix, which is sampled and reshaped into sequences of patches and concatenated with 3D position embeddings. The second part of our model estimates the neighbors of different patches on the basis of learned geometric constraints, and performs spatial propagation leveraging dynamic graph convolution networks with self-attention mechanism. The final depth prediction is obtained through a graph-to-image module. As can be seen in the parts of 3D reconstruction and graph propagation, the neighborhood construction in yellow is varying for different red nodes since the geometric structures surrounding red nodes are diverse in 3D space, and the graphs are also changeable because of the dynamic construction of GCN during the propagation process. Note that the different colors of the edges indicate the different attention weights between the red nodes and their yellow neighbors.}
\label{fig:arch}
\end{figure}

\subsection{GraphCSPN}
After closely examining the updating function and neighborhood construction function of earlier approaches, there are two major limitations that need to be addressed. Firstly, once the affinity matrix $\mathbf{A}_{p,q}$ and neighborhood matrix $\mathcal{N}_{p,q}$ is determined, they will not be changed during the entire process of spatial propagation. So there are only simple iterations and no learning process involved when propagating neighbor observations with corresponding affinities. However, the condition of pixels will be changed after each propagation step, and simple iterations with fixed affinities and neighbors fails to capture the dynamic relationships between pixels. As a result, the fixed configurations slow down the information propagation and demand a large number of iteration steps. And that’s why we develop a dynamically constructed graph convolution network for propagation-efficient depth completion.

The second limitation of previous approaches is that the spatial propagation is confined in a small region and performed in two-dimensional plane without any geometric constraints. Structured kernels (\eg $3\times3$, $5\times5$) used in CSPN~\cite{chen2019learning} and CSPN++~\cite{cheng2020cspn++} have a very limited receptive field for every iteration and can easily bring unrelated information into propagation from fixed neighbors, which may result in inaccurate predictions especially on object boundaries. Although NLSPN~\cite{park2020non} can predict flexible neighbors, the neighbors are learned merely based on the initial depth estimation without explicit geometric constraints, thus leading to unreliable predictions. To fully explore the 3D nature of depth prediction, the model we propose is geometry-aware and capable of capturing relevant neighbors over long distance, which is critical for depth completion because of the irregular distribution of sparse depth measurements. We now explain the process of GraphCSPN in three consecutive steps: graph construction, neighborhood estimation and graph propagation.

\begin{figure*}[t]
\centerline{\includegraphics[width=\linewidth]{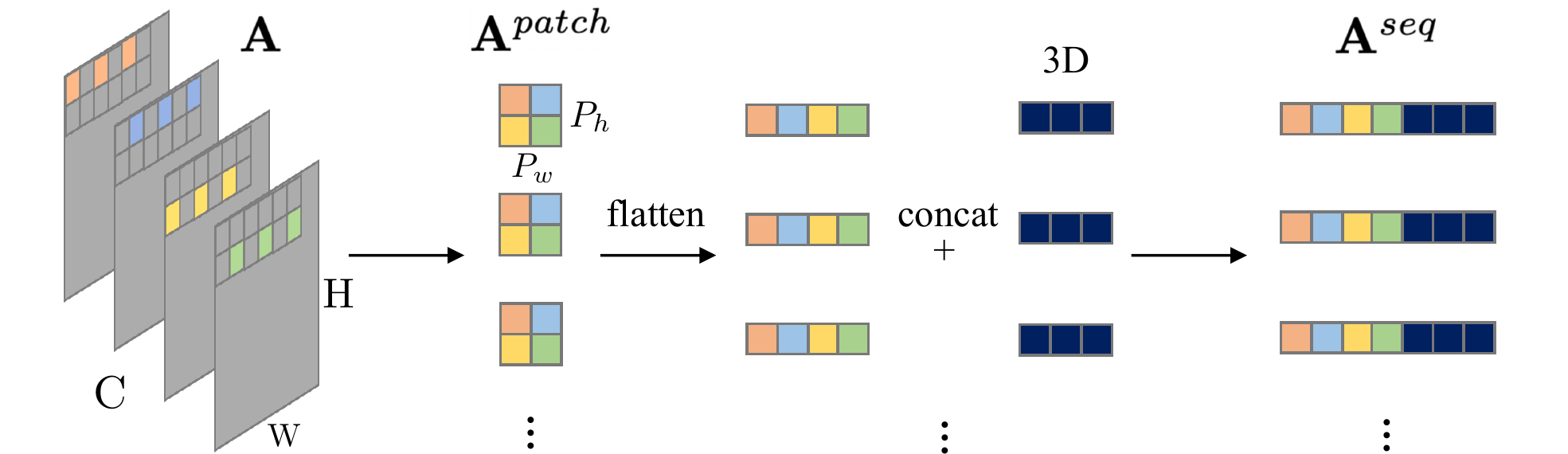}}
\caption{ \textbf{Illustration of the process of graph construction} (Best viewed in color). For better visibility, we set $C=4, P_{h}=P_{w}=2$, and use this simple example to show how to convert affinity maps into a sequence of features.}
\label{fig:gc}
\end{figure*}

{\bf Patch-wise Graph Construction} Graph neural networks take as input a set of node features. So the core of graph construction is to convert affinity maps $\mathbf{A} \in \mathbb{R}^{{H{\times}W{\times}C}}$ into a sequence of features $\mathbf{A}^{seq} \in \mathbb{R}^{{N{\times}L}}$, where $(H, W)$ is the resolution of the affinity map, $C$ is the number of channels, $L$ is the length of node feature, and $N$ is the total number of nodes. Note that the computation complexity of graph propagation increases quadratically with the number of nodes $N$. So we need to keep the size of graph small. To do so, we first convert $\mathbf{A}$ into a sequence of patches $\mathbf{A}^{patch} \in \mathbb{R}^{{N{\times}P_{h}{\times}P_{w}}}$, where $(P_{h}, P_{w})$ is the resolution of each affinity patch. In this way, the total number of nodes $N=(H{\times}W)/(P_{h}{\times}P_{w})$. Otherwise, $N=H{\times}W$, if we use pixel-wise construction (see more discussions in supplement). For each patch $\mathbf{A}^{patch}$ with shape $P_{h}{\times}P_{w}$, the number of corresponding patches in $\mathbf{A}$ is $C$, and the pixels of each patch $\mathbf{A}^{patch}$ is from $C$ different channels ($C=P_{h}{\times}P_{w}$). By our patch-wise construction, prior knowledge about local correlations can be established. And then each patch is flattened and concatenated with 3D position as node features ($L=P_{h}{\times}P_{w}+3$), so the model is also able to capture global relationships of different patches over long distance, as can be seen in the 3D reconstruction part of Fig.\ref{fig:arch}. To help for a better understanding of graph construction, we use a simple example to illustrate the above process in Fig.~\ref{fig:gc}.

{\bf Geometry-aware Neighborhood Estimation} The 3D position embedding preserves the original geometry of the local shape. To explicitly incorporate geometric constraints into the graph propagation, the 3D position embeddings are supposed to be added to patches such that the geometric information can be retained during graph propagation. To do so, we project the depth prediction after each graph propagation step into 3D space as follows:
\begin{equation}
\left[\begin{array}{l}
u \\
v \\
w \\
1
\end{array}\right]=d\left[\begin{array}{cccc}
1 / f_{p} & 0 & -c_{p} / f_{p} & 0 \\
0 & 1 / f_{q} & -c_{q} / f_{q} & 0 \\
0 & 0 & 1 & 0 \\
0 & 0 & 0 & 1
\end{array}\right]\left[\begin{array}{c}
p \\
q \\
1 \\
1 / d
\end{array}\right],
\end{equation}
where $(p,q)$ and $(u,v,w)$ are the center coordinate of the patch in 2D plane and projected 3D space, respectively; $d$ is the depth predictions during propagations; $f_{p},f_{q},c_{p},c_{q}$ are the camera intrinsic parameters. After obtaining the 3D position of each patch, k-nearest neighbors algorithm can be applied for neighborhood estimation in 3D space, which can be formulated as follows:
\begin{equation}
    \mathcal{N}^{G}_{p,q} {=} \{(u,v,w)|(u,v,w) {=} \mathcal{K}(\mathbf{D}| p, q),  (u,v,w) \in \mathbb{R}^{3}\},
\label{eq:myneighbor}
\end{equation}
where $\mathcal{K}$ represents the projection and estimation function; $\mathcal{N}^{G}_{p,q}$ denotes the neighbors of the patch in 3D space. During the propagation process, the graph is dynamically constructed at each propagation step in line with the neighborhood estimation at that time. 

{\bf Dynamic Graph Propagation} After the neighborhood for each patch is constructed, patches are represented as vertices in a graph, and edges can be established if there are connections between the vertices. Graph propagation is conducted via feature aggregation and feature update, which is as follows:
\begin{equation}
\mathbf{h}_{v_{s+1}}=\eta\left(\mathbf{h}_{v_{s}}, \rho\left(\left\{\mathbf{h}_{u_{s}} \mid u_{s} \in \mathcal{N}_{s+1}\left(v_{s}\right)\right\}, \mathbf{h}_{v_{s}}, \mathcal{W}_{\rho}\right), \mathcal{W}_{\eta}\right),
\label{eqn.5}
\end{equation}
where ${s}$ denotes the propagation step as before in Eqn.~\ref{eq:update}; $\rho$ and $\eta$ are the feature aggregation function and update function, $\mathcal{W}_{\rho}$ and $\mathcal{W}_{\eta}$ are the learnable parameters for the two functions, respectively; ${\mathbf{h}_{v_{s}}}$ represents the features for the vertex $v_{s}$; and $\mathcal{N}_{s+1}\left(v_{s}\right)$ is the set of neighbors for $v_{s}$ during the step ${s+1}$. Note that the neighborhood $\mathcal{N}_{s+1}\left(v_{s}\right)$ varies for different propagation steps, since the graph is dynamically constructed. To further facilitate the graph propagation process, we propose a fine-grained edge attention module as follows:
\begin{equation}
\mathbf{x}_{i}^{s+1}=\alpha_{i, i} \phi_{i}(\mathbf{x}_{i}^{s})+\sum_{j \in \mathcal{N}_{s+1}(i)} \alpha_{i, j} \phi_{j}(\mathbf{x}_{i}^{s}||(\mathbf{x}_{j}^{s}-\mathbf{x}_{i}^{s})),
\label{eqn.6}
\end{equation}
where $\mathbf{x}_{i}^{s}$ represents the feature vector for the vertex $i$ at propagtion step $s$; $\|$ denotes the concatenation operation; $\phi$ is a neural network which is instantiated as a multi-layer perceptron (MLP) parameterized by $\mathcal{W}_{\rho}$. The attention is performed in a channel-wise manner and each channel denotes a different position in the original patch, and the attention coefficients $\alpha_{i, j}$ are computed as follows:
\begin{equation}
\alpha_{i, j}=\frac{\exp \left(\psi (\mathbf{x}_{i}||\mathbf{x}_{j})\right)}{\sum_{k \in \mathcal{N}_{s+1}(i) \cup\{i\}} \exp \left(\psi( \mathbf{x}_{i} ||  \mathbf{x}_{k})\right)},
\label{eqn.7}
\end{equation}
where $\psi$ denotes a MLP parameterized by $\mathcal{W}_{\eta}$. Eqn.~\ref{eqn.6} details the computation process of aggregation function $\rho$ and update function $\eta$ stated in Eqn.~\ref{eqn.5}. $\rho$ is instantiated as a MLP $\phi$ which aggregates features of vertices from their neighbors. And $\eta$ works through the attention coefficient $\alpha$ computed by Eqn.~\ref{eqn.7}. By applying the proposed edge attention module, the different attention weights assigned to neighbors can effectively impede the propagation of incorrect predictions during the iterative refinement. After graph propagation, the graph-to-image module rearranges the sequences of patches to their original position in 2D image to generate the final prediction.

\subsection{Overall Architecture}
As can be seen in Figure \ref{fig:arch}, the end-to-end network architecture of our model consists of two parts: an encoder-decoder backbone to learn initial depth prediction and affinity maps, and a graph convolution network for spatial propagation. Follow the common practice of previous work~\cite{cheng2020cspn++,park2020non}, we build the encoder upon residual networks and add mirror connections with the decoder to make up the lost spatial information due to the down sampling and pooling operations in the encoder. Since the input of the encoder-decoder comes from different sensing modalities, depth-wise separable convolution is utilized for the first convolution layer which performs convolution on RGB images and sparse depth map independently, followed by a pointwise convolution as a simple while effective multi-modality fusion strategy. The second part of the model is a geometry-aware graph convolution network equipped with edge attention, which is dynamically constructed for iterative refinement of depth predictions.

\subsection{Loss Function}
The model is trained with masked $\ell_{1}$ loss between the ground truth depth map and predicted depth map, which is the same loss function used in previous methods~\cite{ma2018sparse,cheng2018depth,park2020non} and defined as follows: 
\begin{equation}
    \mathcal{L}(\mathbf{D}^{gt}, \mathbf{D}^{pred}) = \frac{1}{N_{v}} \sum_{i,j}\mathbb{I}(d_{i,j}^{gt}>0)\left\vert~d_{i,j}^{gt} - d_{i,j}^{pred}~\right\vert,
\label{eq:loss}
\end{equation}
where $\mathbb{I}$ is the indication function; $d_{i,j}^{gt}\in\mathbf{D}^{gt}$ and $d_{i,j}^{pred}\in\mathbf{D}^{pred}$ represent the ground truth depth map and the depth map predicted by our model; and ${N_{v}}$ denotes the total number of valid depth pixels. We have explored other loss functions, such as $\ell_{2}$ loss and smooth $\ell_{1}$ loss, and also find that the training process can be accelerated by adding extra supervision as weighted auxiliary loss to the output of each graph propagation. More details about the setting of loss functions can be found in ablation.

\section{Experiments}
We conduct extensive experiments and ablation studies to verify the effectiveness of our model. In this section, we first briefly introduce the datasets and evaluation metrics. Then we present qualitative and quantitative comparisons with state-of-the-arts, and ablation studies of each component of our model are also provided. Additional experiment details can be found in supplementary materials.

\begin{figure*}[t]
\begin{center}
\renewcommand{\arraystretch}{0.2}
\begin{tabular}{@{}c@{\hskip 0.001\linewidth}c@{\hskip 0.001\linewidth}c@{\hskip 0.001\linewidth}c@{\hskip 0.001\linewidth}c@{\hskip 0.001\linewidth}c@{\hskip 0.001\linewidth}c}
\includegraphics[width=0.14\linewidth]{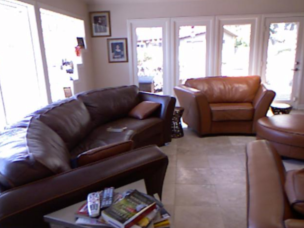} &
\includegraphics[width=0.14\linewidth]{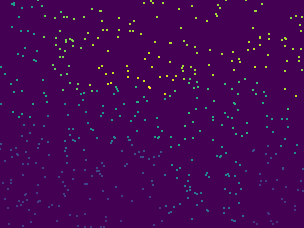} &
\includegraphics[width=0.14\linewidth]{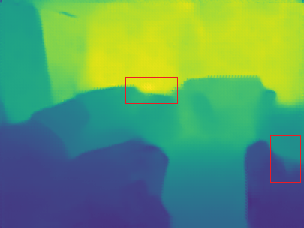} &
\includegraphics[width=0.14\linewidth]{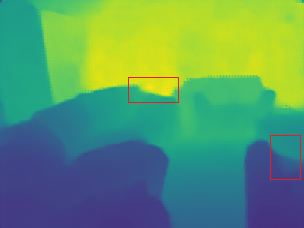} &
\includegraphics[width=0.14\linewidth]{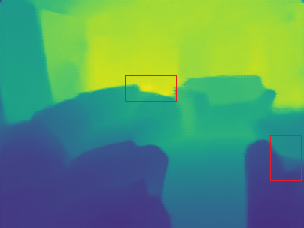} &
\includegraphics[width=0.14\linewidth]{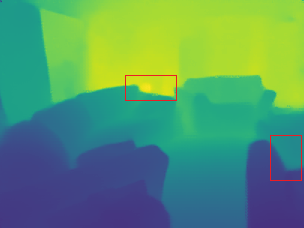} &
\includegraphics[width=0.14\linewidth]{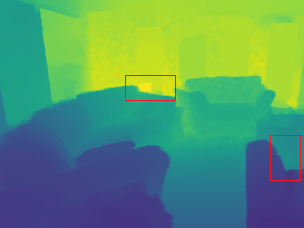} \\
\includegraphics[width=0.14\linewidth]{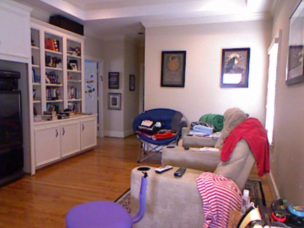} &
\includegraphics[width=0.14\linewidth]{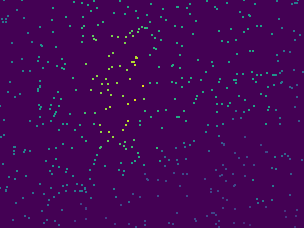} &
\includegraphics[width=0.14\linewidth]{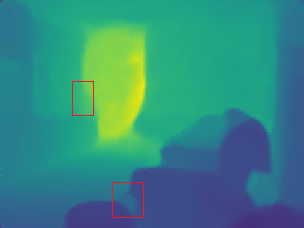} &
\includegraphics[width=0.14\linewidth]{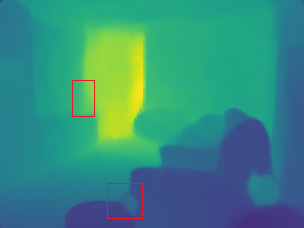} &
\includegraphics[width=0.14\linewidth]{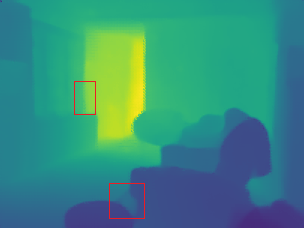} &
\includegraphics[width=0.14\linewidth]{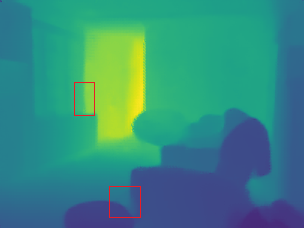} &
\includegraphics[width=0.14\linewidth]{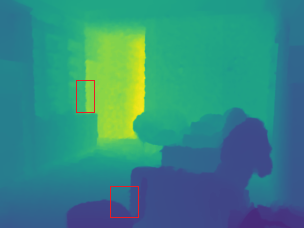} \\
{ \scriptsize(a) RGB} & { \scriptsize(b) Depth} & { \scriptsize(c) SPN} &
{\scriptsize(d) CSPN} & { \scriptsize(e) NLSPN} & { \scriptsize(f) Ours} &
{ \scriptsize(g) GT}
\end{tabular}
\caption{\textbf{Depth completion results on the NYU-Depth-v2 dataset~\cite{silberman2012indoor}}.The area inside the red bounding box reveals that our model recovers more accurate structures. (Better viewed in color and zoom in.)}
\label{fig:result_nyu}
\end{center}
\end{figure*}

\subsection{Datasets and Metrics}
{\bf NYU-Depth-v2} The NYU-Depth-v2 dataset~\cite{silberman2012indoor} provides RGB and depth images of 464 different indoor scenes with the Kinect sensor. We use the official split of data, where 249 scenes are used for training and the remaining 215 scenes for testing. And we sample about 50K images out of the training set and the original frames of size 640×480 are first downsampled to half and then 304×228 center-cropping is applied, as seen in previous work~\cite{laina2016deeper,eigen2014depth,ma2018sparse}. Following the standard setting~\cite{ma2018sparse,cheng2018depth}, the official test dataset with 654 images is used for evaluating the final performance. 

\noindent{\bf KITTI Dataset} KITTI dataset~\cite{geiger2012we,geiger2013vision} is a large self-driving real-world dataset with over 90k paired RGB images and LiDAR depth measurements. We use two versions of KITTI dataset, and one is from~\cite{ma2018sparse}, which consists of 46k images from the training sequences for training, and a random subset of 3200 images from the test sequences for evaluation. The other is KITTI Depth Completion dataset~\cite{uhrig2017sparsity} which provides 86k training, 7k validation and 1k testing depth maps with corresponding raw LiDAR scans and reference images. And we evaluate and test the model on the official selected validation set and test set.

\noindent{\bf Metrics} We adopt the same metrics used in previous work~\cite{ma2018sparse,cheng2018depth,uhrig2017sparsity} on NYU-Depth-v2 dataset and KITTI dataset. Given ground truth depth map $D^* = \{d^*\}$ and predicted depth map $D = \{d\}$, the metrics used in the following experiments include:

(1)RMSE:$\sqrt{\frac{1}{|D|}\sum_{d \in D}|d^* - d|^2}$;

(2)MAE: $\frac{1}{|D|}\sum_{d \in D}|d^* - d|$;

(3)iRMSE: $\sqrt{\frac{1}{|D|}\sum_{d \in D}|1/d^* - 1/d|^2}$;

(4)iMAE: $\frac{1}{|D|}\sum_{d \in D}|1/d^* - 1/d|$;

(5)REL: $\frac{1}{|D|}\sum_{d \in D}|d^* - d|/d^*$;

(6)$\delta_t$: percentage of pixels satisfying $max(\frac{d^*}{d}, \frac{d}{d^*})<t$, where $t \in \{1.25, 1.25^2, 1.25^3\}$.

\begin{figure*}[t]
\begin{center}
\renewcommand{\arraystretch}{0.2}
\begin{tabular}{@{}c@{\hskip 0.001\linewidth}c@{\hskip 0.001\linewidth}c@{\hskip 0.001\linewidth}c}
\raisebox{2\height}{{\csmall (a)}}  &
\includegraphics[width=0.32\linewidth]{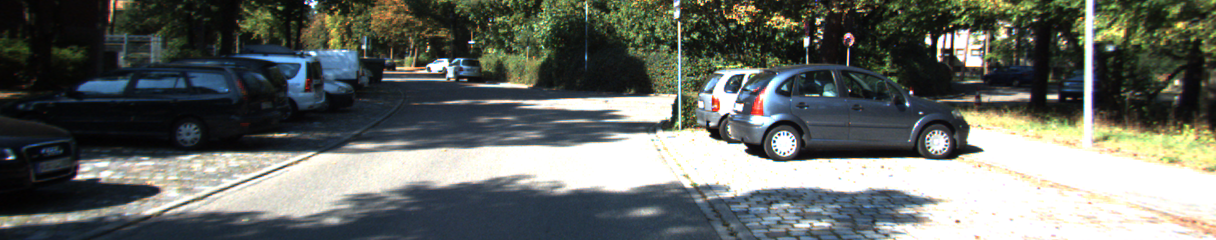} &
\includegraphics[width=0.32\linewidth]{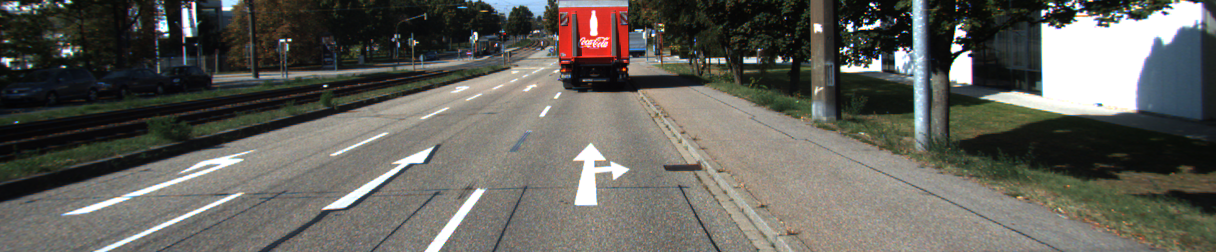} &
\includegraphics[width=0.32\linewidth]{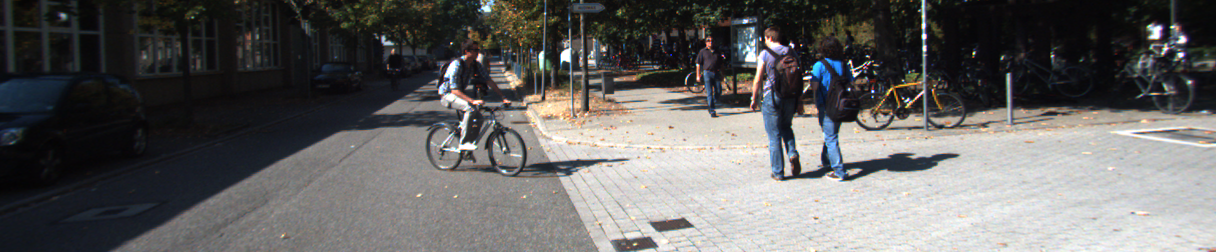} \\
\raisebox{2\height}{{\csmall (b)}}  &
\includegraphics[width=0.32\linewidth]{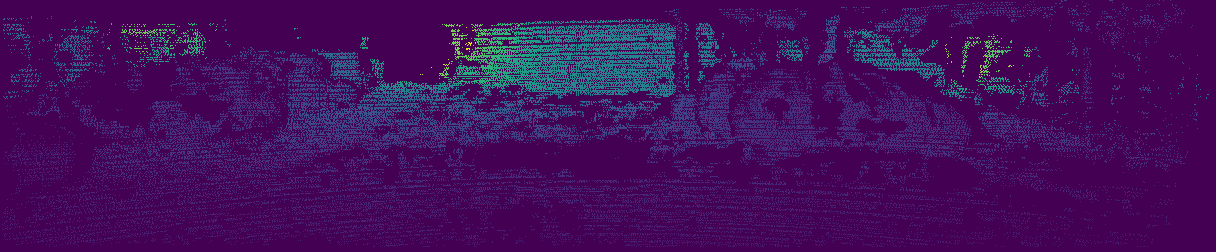} &
\includegraphics[width=0.32\linewidth]{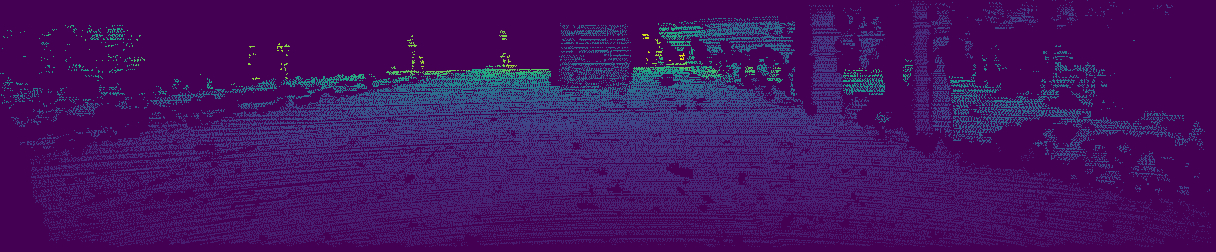} &
\includegraphics[width=0.32\linewidth]{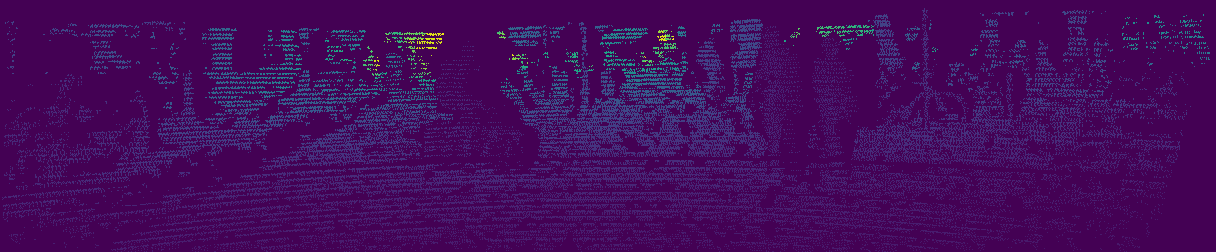} \\
\raisebox{2\height}{{\csmall (c)}}  &
\includegraphics[width=0.32\linewidth]{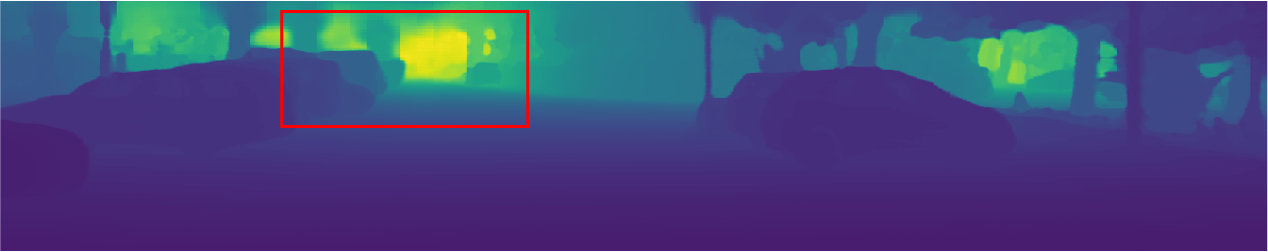} &
\includegraphics[width=0.32\linewidth]{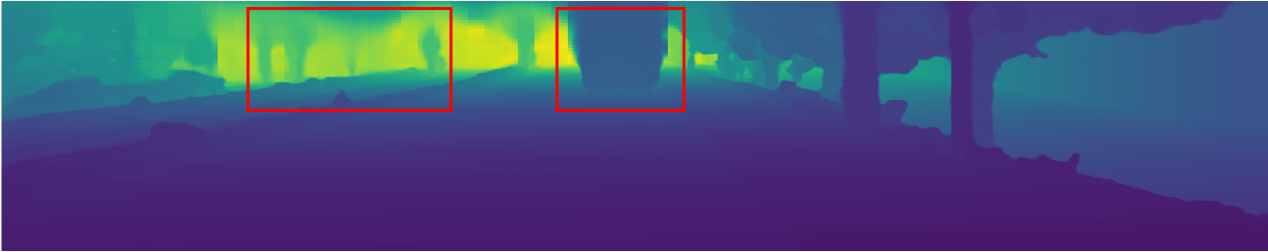} &
\includegraphics[width=0.32\linewidth]{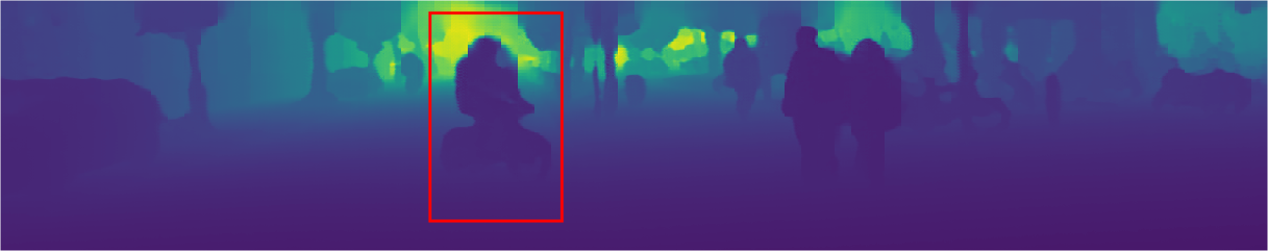} \\
\raisebox{2\height}{{\csmall (d)}}  &
\includegraphics[width=0.32\linewidth]{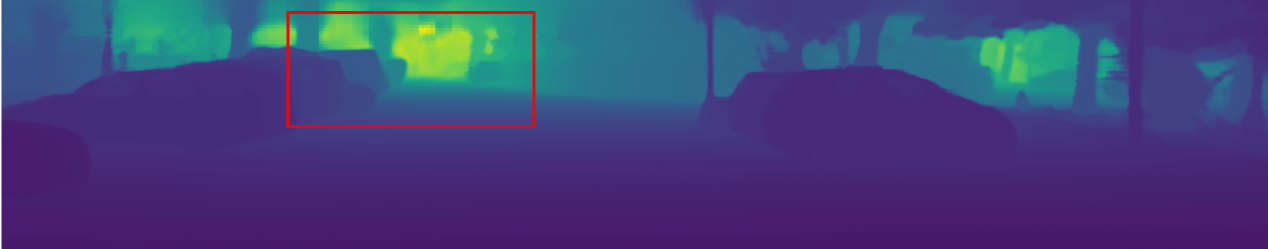} &
\includegraphics[width=0.32\linewidth]{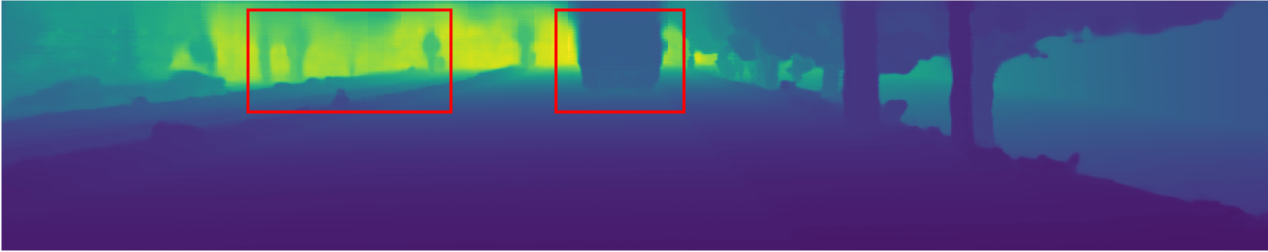} &
\includegraphics[width=0.32\linewidth]{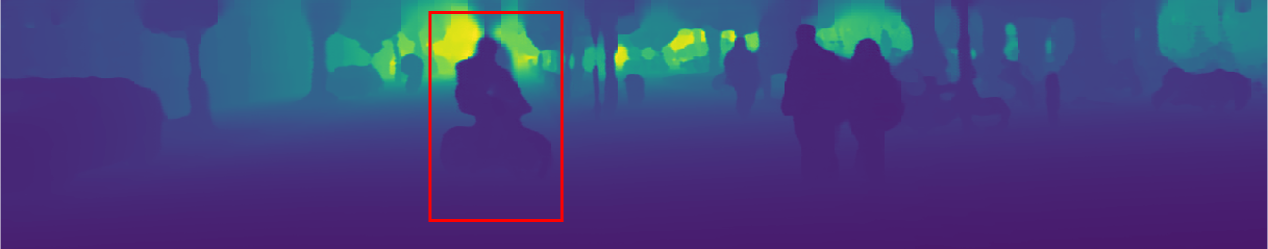} \\
\raisebox{2\height}{{\csmall (e)}}  &
\includegraphics[width=0.32\linewidth]{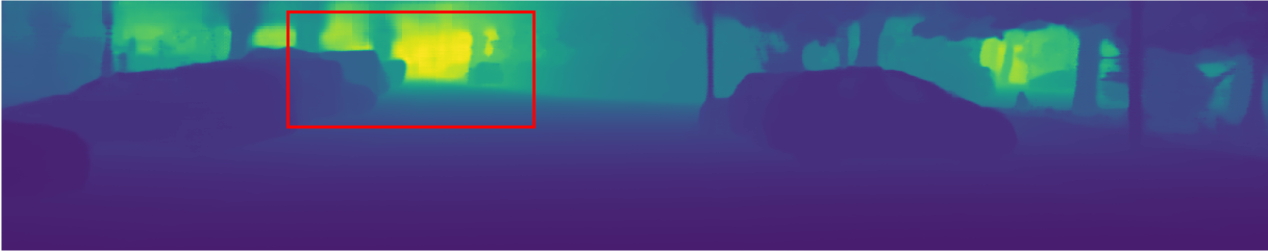} &
\includegraphics[width=0.32\linewidth]{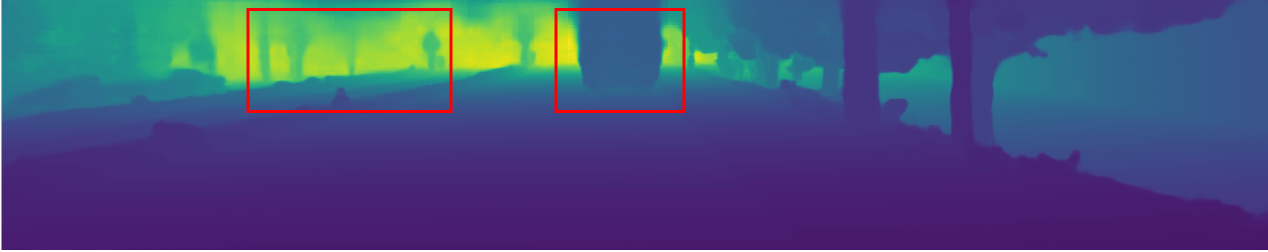} &
\includegraphics[width=0.32\linewidth]{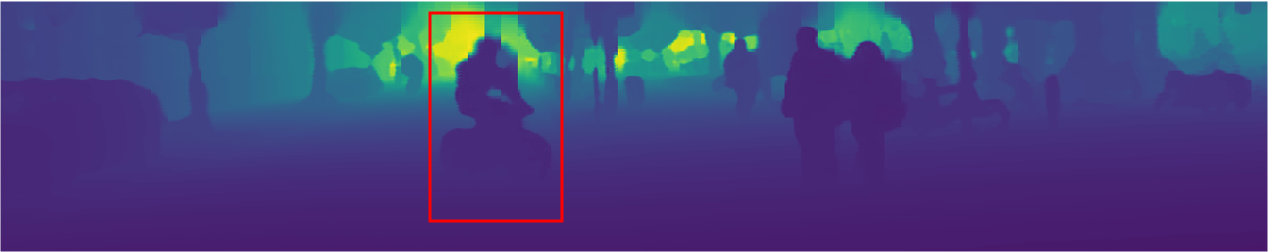} \\
\raisebox{2\height}{{\csmall (f)}}  &
\includegraphics[width=0.32\linewidth]{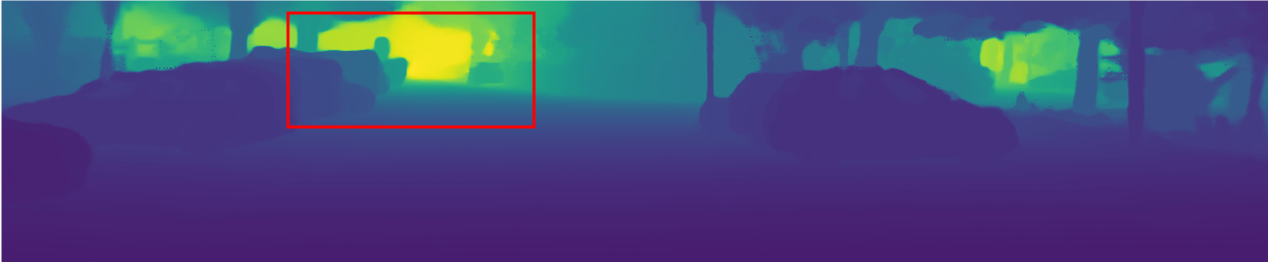} &
\includegraphics[width=0.32\linewidth]{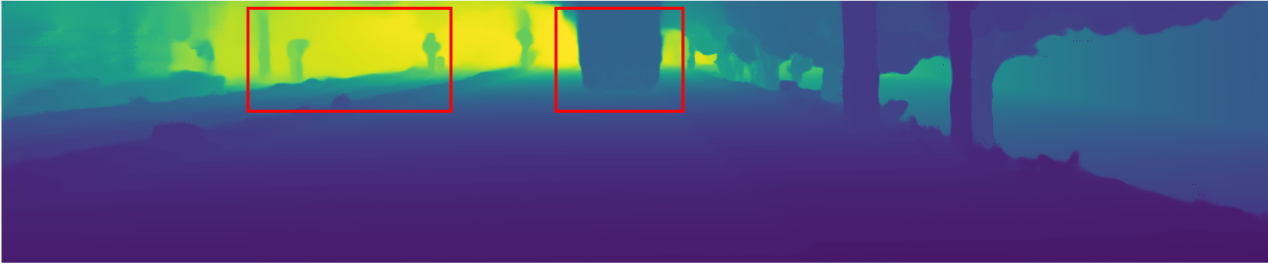} &
\includegraphics[width=0.32\linewidth]{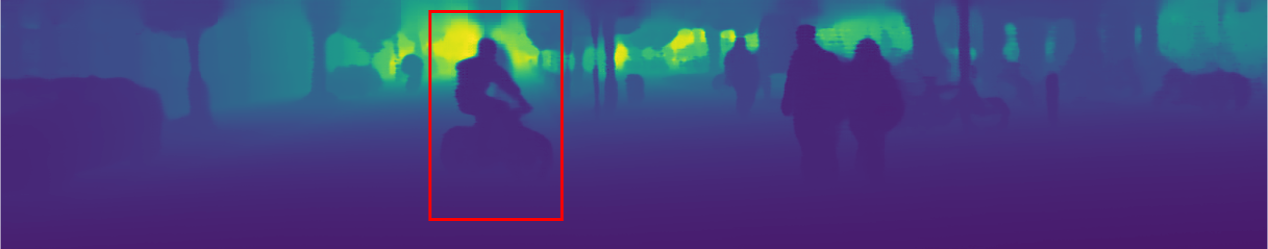} \\
\end{tabular}
\caption{\textbf{Depth completion results on the KITTI Depth Completion dataset~\cite{uhrig2017sparsity}}. (a) RGB, (b) Ground truth, (c) CSPN~\cite{cheng2018depth}, (d) DepthNormal~\cite{xu2019depth}, (e) NLSPN~\cite{park2020non}, (f) Ours. (Better viewed in color and zoom in.)}
\label{fig:result_kittidc}
\end{center}
\end{figure*}

\subsection{Comparison with State-of-the-arts}
We evaluate our model on the official test split of NYU-Depth-v2 dataset and 500 depth pixels are randomly sampled from the dense depth map, as in previous works~\cite{ma2018sparse,cheng2018depth,park2020non}. Quantitative comparison results with methods based on spatial propagation networks are provided in Table~\ref{tab:nyu}. S2D~\cite{ma2018sparse} is a direct depth completion model which can only obtain blurry results. And spatial propagation networks based approaches are able to generate much more accurate and reliable results by recurrently refining the predictions. However, earlier methods have not taken the geometric constraints into account and fail to capture global relationship over long distance, thus easily leading to unreliable results on object boundaries where a holistic view is need, as can be seen in Figure~\ref{fig:result_nyu}.
Moreover, the affinity matrix and neighborhood of previous methods are fixed during propagation, and a large number of iteration steps is required because of the inefficient propagation. The results in Table~\ref{tab:nyu} show that our model achieves significant improvements over previous approaches when compared using only three iteration steps.

\begin{table}[t]
\centering
\footnotesize
\setlength\tabcolsep{6pt} 

\begin{tabular}{| c | c || *{2}{ c } | *{3}{ c } | }
\hline
Methods & Iterations & RMSE $\downarrow$ & REL $\downarrow$ & $\delta_{1.25}$ $\uparrow$ & $\delta_{1.25^{2}}$ $\uparrow$ & $\delta_{1.25^{3}}$ $\uparrow$\\ \hline\hline 

S2D~\cite{ma2018sparse}
& - & 0.230 & 0.044 & 97.1 & 99.4 & 99.8 \\ 
\hline 

SPN~\cite{liu2017learning}
& 3 & 0.215 & 0.040 & 94.2 & 97.6 & 98.9 \\
& 24 & 0.172 & 0.031 & 98.3 & 99.7 & 99.9 \\
\hline 

CSPN~\cite{cheng2018depth}
& 3 & 0.135 & 0.021 & 97.1 & 98.1 & 99.3 \\
& 24 & 0.117 & 0.016 & 99.2 & 99.9 & 100.0 \\
\hline 

CSPN++~\cite{cheng2020cspn++}
& 24 & 0.116 & - & - & - & - \\
\hline 

NLSPN~\cite{park2020non}
& 3 & 0.119 & 0.018 & 98.3 & 99.4 & 99.3 \\
& 18 & 0.092 & 0.012 & 99.6 & 99.9 & 100.0 \\
\hline 

GraphCSPN
& \bf 3 & \bf 0.090 & \bf 0.012 & \bf 99.6 & \bf 99.9 & \bf 100.0 \\
\hline

\end{tabular}

\caption{Quantitative evaluation on the NYU-Depth-v2 dataset. We highlight the best results when propagating for only 3 steps.}

\label{tab:nyu}
\end{table}

\begin{table}[t]
\centering
\footnotesize
\setlength\tabcolsep{6pt} 

\begin{tabular}{| c | c || *{2}{ c } | *{3}{ c } | }
\hline
Methods & Iterations & RMSE $\downarrow$ & REL $\downarrow$ & $\delta_{1.25}$ $\uparrow$ & $\delta_{1.25^{2}}$ $\uparrow$ & $\delta_{1.25^{3}}$ $\uparrow$\\ \hline\hline 

S2D~\cite{ma2018sparse}
& - & 3.378 & 0.073 & 93.5 & 97.6 & 98.9 \\ 
\hline 

SPN~\cite{liu2017learning}
& 3 & 3.302 & 0.069 & 93.7 & 97.6 & 99.0 \\
& 24 & 3.243 & 0.063 & 94.3 & 97.8 & 99.1 \\
\hline 

CSPN~\cite{cheng2018depth}
& 3 & 3.125 & 0.052 & 94.1 & 97.8 & 99.0 \\
& 24 & 2.977 & 0.044 & 95.7 & 98.0 & 99.1 \\
\hline 

NLSPN~\cite{park2020non}
& 3 & 2.697 & 0.042 & 95.8 & 98.0 & 99.1 \\
& 18 & 2.533 & 0.038 & 96.2 & 98.5 & 99.3 \\
\hline 

GraphCSPN
& \bf 3 & \bf 2.267 & \bf 0.032 & \bf 97.4 & \bf 98.9 & \bf 99.5 \\
\hline

\end{tabular}

\caption{Quantitative evaluation on the KITTI dataset.}

\label{tab:kitti}
\end{table}

The Table~\ref{tab:kitti} shows the comparisons of evaluation results on KITTI dataset~\cite{ma2018sparse}. Similar to the result on NYU-Depth-v2 dataset, our model outperforms the previous methods by a large margin which reduces 0.4m in RMSE. And our model also achieves better results compared to other geometry-aware methods~\cite{imran2019depth,qiu2019deeplidar,xu2019depth} and other methods using GNNs~\cite{xiong2020sparse,zhao2020adaptive}, as can be seen in Table~\ref{tab:nyu2}. On the challenging KITTI Depth Completion dataset~\cite{uhrig2017sparsity} where the ground truth of depth map is also sparse. Our model still attains competitive results compared to other state-of-the-arts, as can be seen in Table~\ref{tab:kitti2}. Qualitative results of the proposed method in comparison with state-of-the-arts are shown in Figure~\ref{fig:result_kittidc}. We also provide a video demo of our method using this dataset in supplement.

\begin{table}[t]
{\scriptsize
    \begin{minipage}[t]{0.45\linewidth}
        \begin{center}
        \renewcommand{\arraystretch}{1.4}
        \setlength\tabcolsep{2pt} 
        \begin{tabular}{c|c c c c }
        \hline
        Method & RMSE & REL & $\delta_{1.25}$ & $\delta_{1.25^{2}} $ \\ \hline 
        DepthCoeff~\cite{imran2019depth} & 0.118 & 0.013 & 99.4 & 99.9  \\ 
        DeepLiDAR~\cite{qiu2019deeplidar} & 0.115 & 0.022 & 99.3 & 99.9 \\ 
        DepthNormal~\cite{xu2019depth} & 0.112 & 0.018 & 99.5 & 99.9 \\ 
        GNN~\cite{xiong2020sparse} & 0.106 & 0.016 & 99.5 & 99.9 \\ 
        FCFRNet~\cite{liu2021fcfr} & 0.106 & 0.015 & 99.5 & 99.9 \\ 
        ACMNet~\cite{zhao2020adaptive} & 0.105 & 0.015 & 99.4 & 99.9 \\ 
        PRNet~\cite{lee2021depth} & 0.104 & 0.014 & 99.4 & 99.9 \\ 
        GuideNet~\cite{tang2020learning} & 0.101 & 0.015 & 99.5 & 99.9 \\ 
        TWICE~\cite{imran2021depth} & 0.097 & 0.013 & 99.6 & 99.9 \\ 
        GraphCSPN & \bf 0.090 & \bf 0.012 & \bf 99.6 & \bf 99.9  \\ \hline
        \end{tabular}

\caption{Comparison on NYU-Depth-v2 dataset with other state-of-the-arts. }

\label{tab:nyu2}
\end{center}
    \end{minipage}
    \hfill
     \begin{minipage}[t]{0.49\linewidth}
        \begin{center}
        \renewcommand{\arraystretch}{1.4}
        \setlength\tabcolsep{2pt} 
        \begin{tabular}{c|c c c c c}
        \hline
        Method & RMSE & MAE & iRMSE & iMAE \\ \hline 
        CSPN~\cite{cheng2018depth} & 1019.64 & 279.46 & 2.93 & 1.15 \\ 
        TWICE~\cite{imran2021depth} & 840.20 & \textbf{195.58} & 2.08 & \textbf{0.82} \\
        DepthNormal~\cite{xu2019depth} & 777.05 & 235.17 & 2.42 & 1.13 \\ 
        DeepLiDAR~\cite{qiu2019deeplidar} & 758.38 & 226.50 & 2.56 & 1.15 \\ 
        CSPN++~\cite{cheng2020cspn++} & 743.69 & 209.28 & 2.07 & 0.90 \\ 
        NLSPN~\cite{park2020non} & 741.68 & 199.59 & 1.99 & 0.84 \\ 
        GuideNet~\cite{tang2020learning} & 736.24 & 218.83 & 2.25 & 0.99 \\
        FCFRNet~\cite{liu2021fcfr} & 735.81 & 217.15 & 2.20 & 0.98 \\
        PENet~\cite{hu2021penet} & \textbf{730.08} & 210.55 & 2.17 & 0.94 \\
        GraphCSPN & 738.41 & 199.31 & \textbf{1.96} & 0.84 \\ 
        \hline
        \end{tabular}

\caption{Comparison on KITTI Depth Completion test dataset.}

\label{tab:kitti2}
        \end{center}
    \end{minipage}
}
\end{table}

\subsection{Ablation Studies}
We conduct extensive experiments to verify the effectiveness of each proposed component. The altered components of our model include: iteration steps, the number of neighbors, the sparsity of depth samples, loss functions, the proposed edge attention and geometry-aware modules, and dynamic graph construction.

\begin{figure}[t]
\centering
\includegraphics[width=\linewidth]{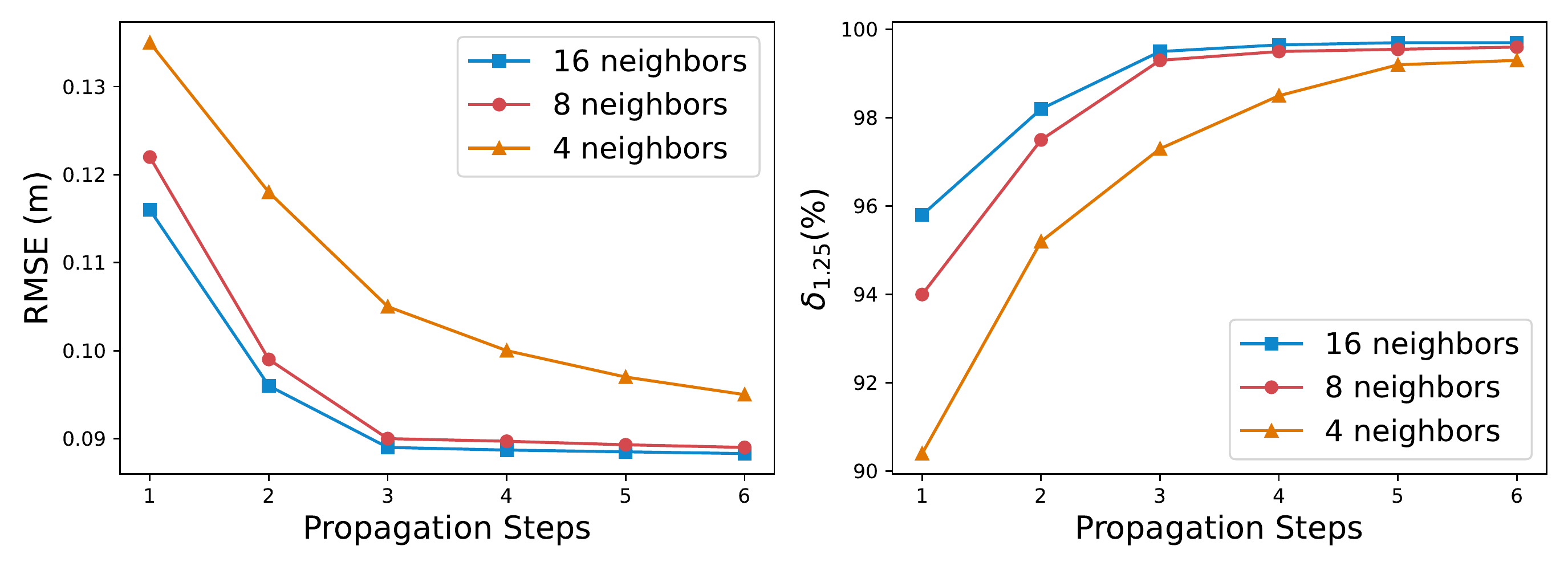} 
\caption{Impact of number of propagation steps and neighbors on the prediction accuracy on NYU-Depth-v2 dataset.}
\label{fig:ab_nyu}
\end{figure}

{\bf Propagation Configurations:} There are two important factors in our proposed graph convolution based spatial propagation network, which are the iteration steps and the number of neighbors. To investigate the impact of those factors on final results, we set the iteration steps from 1 to 6 and the number of neighbors to 4, 8, 16. The results in Figure~\ref{fig:ab_nyu} indicate that it is difficult to aggregate enough information when propagating for only one step, so the results get worse. And when propagating for six steps which is larger than our original model, the model only achieves slightly better results, which implies that a small number of iteration steps is already adequate since our model is propagation-efficient. As for the size of neighborhood, we find that a larger number of neighbors can slightly improve the results while the results degrade when there are only limited neighbors, which also takes more epochs to converge. So a combination of larger number of propagation steps and neighbors can achieves better performance which demonstrate the capacity and potential of our model, but we choose a small number instead for a balance of accuracy and efficiency. Since GNN can directly work on the irregular input data, we also try to apply the graph propagation to the raw input. The results in Table~\ref{tab:input_ini} show it needs more steps to converge and leads to poor performance. The learned initial depth map can be viewed as a good starting point for a fast optimization process. Moreover, the encoder-decoder architecture to generate the initial depth is an effective fusion network to learn the joint representation of multi-modality RGB and depth images. And it can work in a complementary way with GNN.

{\bf Loss Functions:} Our model is driven by $\ell_{1}$ loss. Here we use $\ell_{2}$ loss and smooth $\ell_{1}$ loss to study the effect of different loss functions, as they play a critical role in model training. Although it is expected that a better result in RMSE would be obtained when applying $\ell_{2}$ loss, in experiments we do not see the expected gains and find that it takes more epochs to reach results close to $\ell_{1}$ loss, because $\ell_{2}$ loss is sensitive to outliers and results in very small gradients during the late stage of training. Smooth $\ell_{1}$ loss is a compromise between $\ell_{2}$ loss and $\ell_{1}$ loss, but $\ell_{1}$ loss is a more robust and suitable choice for the task of depth prediction. We also examine the impact of auxillary loss on model training and final results by adding extra supervision to the intermediate outputs of each graph propagation with weight 0.1. The results in Table~\ref{tab:loss} show that model training benefits from the auxillary loss with a faster speed of convergence, because the intermediate outputs share the same objective with the final output.

\begin{table}[t]
{\small
    \begin{minipage}[t]{0.44\linewidth}
        \begin{center}
        \setlength\tabcolsep{2pt} 
        \begin{tabular}{l|c|c|c}
\hline
loss & epochs & RMSE & REL \\
\hline
$\ell_1$ & 40 & 0.094 & 0.013\\
smooth $\ell_1$ & 42 & 0.102 & 0.014\\
$\ell_2$ & 46 & 0.104 & 0.015\\
\hline
auxillary loss & 36 & 0.090 & 0.012\\
\hline
\end{tabular}
\caption{Ablation study on the choices of loss function.}
\label{tab:loss}
\end{center}
    \end{minipage}
    \hfill
    \begin{minipage}[t]{0.54\linewidth}
        \begin{center}
        \setlength\tabcolsep{2pt} 
\begin{tabular}{c|c|c|c|c}
\hline
attention & geometry & dynamic & RMSE & REL \\
\hline
& \checkmark & \checkmark & 0.101 & 0.014  \\ 
\checkmark &  & \checkmark & 0.113 & 0.017  \\
\checkmark & \checkmark &  & 0.108 & 0.016  \\
\checkmark& \checkmark & \checkmark & 0.090 & 0.012 \\
\hline
\end{tabular}
\caption{Ablation study on the configurations of graph construction.}
\label{tab:ablation_graph}
  \end{center}
    \end{minipage}
}
\end{table}

\begin{table}[t]
{\small
    \begin{minipage}[t]{0.43\linewidth}
        \begin{center}
        \setlength\tabcolsep{3pt} 
        \begin{tabular}{l|c|c|c}
\hline
initialization & epochs & RMSE & REL\\
\hline
raw input & 50 & 0.134 & 0.022 \\
\hline
\end{tabular}
\caption{Ablation study on the initialization of graph propagation.}
\label{tab:input_ini}
\end{center}
    \end{minipage}
    \hfill
    \begin{minipage}[t]{0.53\linewidth}
        \begin{center}
        \setlength\tabcolsep{4pt} 
\begin{tabular}{l|c|c|c|c|c}
\hline
sparsity & 200 & 400 & 500 & 600 & 800 \\
\hline
RMSE & 0.119 & 0.104 & 0.090 & 0.087 & 0.082 \\
\hline
\end{tabular}
\caption{Ablation study on the number of sparse depth samples.}
\label{tab:input}
  \end{center}
    \end{minipage}
}
\end{table}

{\bf Geometry, Attention and Dynamic Construction:} In this part, we verify the role of each component of our graph propagation including the geometry-aware module, edge attention, and dynamic construction. Our model explicitly incorporates geometric constraints into the process of propagation. After removing those constraints, the neighborhood estimation can only perform in the feature space of patches with no access to the knowledge of their real locations in 3D space. As a result, the performance drops by a large margin which implies the importance of 3D geometric clues in the task of depth completion. We further remove the edge attention module from our framework and use the mean aggregation function instead to verify the necessity of attention-driven aggregation. As shown in Table~\ref{tab:ablation_graph}, the performance of the model without edge attention also decreases with a small degree, because the edge attention module can effectively impede the propagation of errors in refinement module. In addition, when the graph propagation is not constructed dynamically and shares the same neighborhood estimation during all the propagation steps, we can see the performance gets worse and converges at an earlier epoch. Because there is an inherent over-smoothing problem of vanilla graph convolution networks, and the dynamic construction of graph is capable of preventing such problem and helps to achieve more accurate results.

{\bf Robustness and Generalization:} Following the standard procedure, our original model works on sparse depth map with 500 random samples. To evaluate the robustness of our model to different input sparsity, we test our model using sparse depth map with 200 samples which takes only 0.4\% of the dense depth map. Although the performance decreases as expected, the model can still generate reasonable results and when changing the sparse input to 800 samples, the model attains a result of 0.083m evaluated by RMSE metric, which demonstrates our model is robust and generalizes well to different input sparsity. Please refer to the supplement for additional ablation studies and visualizations.

\section{Conclusion}
In this paper, we have proposed a graph convolution based spatial propagation network for sparse-to-dense depth completion. The proposed method generalizes previous spatial propagation based approaches into a unified framework which is geometry-aware and propagation-efficient. The graph propagation is dynamically constructed and performed with an edge attention module for feature aggregation and update. Extensive experiments demonstrate the effectiveness of the proposed method. Since our model is a generic and effective solution for 3D spatial propagation, it can be further extended to more 3D perception related tasks in the future.

\noindent\textbf{Acknowledgement.} This work was supported by the state key development program in 14th Five-Year under Grant Nos.2021QY1702, 2021YFF0602103, 2021YFF0602102. We also thank for the research fund under Grant No. 2019GQ\\G0001 from the Institute for Guo Qiang, Tsinghua University.

%
%
\bibliographystyle{splncs04}
\bibliography{egbib}
\end{document}